\documentclass[letterpaper, 10 pt, conference]{ieeeconf} 

\usepackage[dvipsnames]{xcolor}
\usepackage{graphicx, makecell, algorithm, algpseudocode, color, tcolorbox, adjustbox, float, multicol}

\IEEEoverridecommandlockouts                              

\usepackage{graphics} 
\usepackage{times} 
\usepackage{amsmath} 
\usepackage{amssymb}  
\usepackage{mathrsfs}
\usepackage{amsfonts}
\usepackage{kantlipsum}
\usepackage{subfigure}

\makeatletter
\let\NAT@parse\undefined
\makeatother
\usepackage[numbers,sort&compress]{natbib}

\usepackage{hyperref}
\usepackage{graphicx}
\usepackage{float}
\usepackage{ctable}
\usepackage{cuted}
\usepackage{colortbl}
\definecolor{Gray}{gray}{0.9}
\definecolor{White}{gray}{1}
\usepackage{multirow}
\usepackage[font=small,labelfont=bf,tableposition=top]{caption}
\usepackage{listings}

\usepackage{framed}

\newcommand{\tabincell}[2]{\begin{tabular}{@{}#1@{}}#2\end{tabular}} 
\newlength\savewidth

\usepackage{array}
\newcolumntype{?}{!{\vrule width 0.7pt}}
\let\labelindent\relax
\usepackage{enumitem}
\setlist[itemize]{leftmargin=3mm}

\makeatletter
\newcommand\notsotiny{\@setfontsize\notsotiny{6.11415}{7.1828}}
\makeatother

\title{\LARGE \bf
Generalizable Long-Horizon Manipulations\\ with Large Language Models 

}

\author{Haoyu Zhou$^{1}$, Mingyu Ding$^{2}$$^{*}$, Weikun Peng$^{1}$, Masayoshi Tomizuka$^{2}$, Lin Shao$^{1}$$^{*}$ and Chuang Gan$^{3,4}$
\thanks{$^{1}$National University of Singapore
        {\tt\small zhouhaoyu01@u.nus.edu, linshao@nus.edu.sg}
        }%
\thanks{$^{2}$UC Berkeley, USA
        {\tt\small \{myding, tomizuka\}@berkeley.edu}
        }
\thanks{$^{3}$University of Massachusetts Amherst, USA}
\thanks{$^{4}$MIT-IBM Watson AI Lab, USA {\tt\small{ganchuang1990@gmail.com}}}
\thanks{$^{*}$Corresponding authors}
}

\begin{document}
\maketitle{}
\thispagestyle{empty}
\pagestyle{empty}

\begin{abstract}
This work introduces a framework harnessing the capabilities of Large Language Models (LLMs) to generate primitive task conditions for generalizable long-horizon manipulations with novel objects and unseen tasks.
These task conditions serve as guides for the generation and adjustment of Dynamic Movement Primitives (DMP) trajectories for long-horizon task execution.  
We further create a challenging robotic manipulation task suite based on Pybullet for long-horizon task evaluation.
Extensive experiments in both simulated and real-world environments demonstrate the effectiveness of our framework on both familiar tasks involving new objects and novel but related tasks, highlighting the potential of LLMs in enhancing robotic system versatility and adaptability.
Project website: \url{https://object814.github.io/Task-Condition-With-LLM/}.


\end{abstract}


\section{Introduction} \label{sec: introduction}

Recent years have witnessed significant achievements in robot manipulations, and the growing demand for household and multifunctional robots capable of handling complex tasks has brought long-horizon manipulations into the spotlight.

Approaches like Task and Motion Planning~\cite{kaelbling2011hierarchical,migimatsu2020object,toussaint2015logic} and hierarchical reinforcement/imitation learning methods~\cite{barto2003recent,kulkarni2016hierarchical} propose to decompose long-horizon tasks into hierarchies, comprising high-level primitive tasks or discrete symbolic states alongside low-level manipulation motions.
However, the full promise of long-horizon tasks necessitates not only task decomposition with primitive tasks but also a keen focus on environmental conditions. These conditions encompass aspects such as object interactions and spatial relationships, \emph{e.g.}, an object inside another or a gripper grasping an object, playing an essential role in determining the success or failure of primitive tasks. They are crucial for guiding and correcting low-level trajectories and motions during long-horizon execution.

\begin{figure}[t]
    \includegraphics[width=0.99\linewidth]{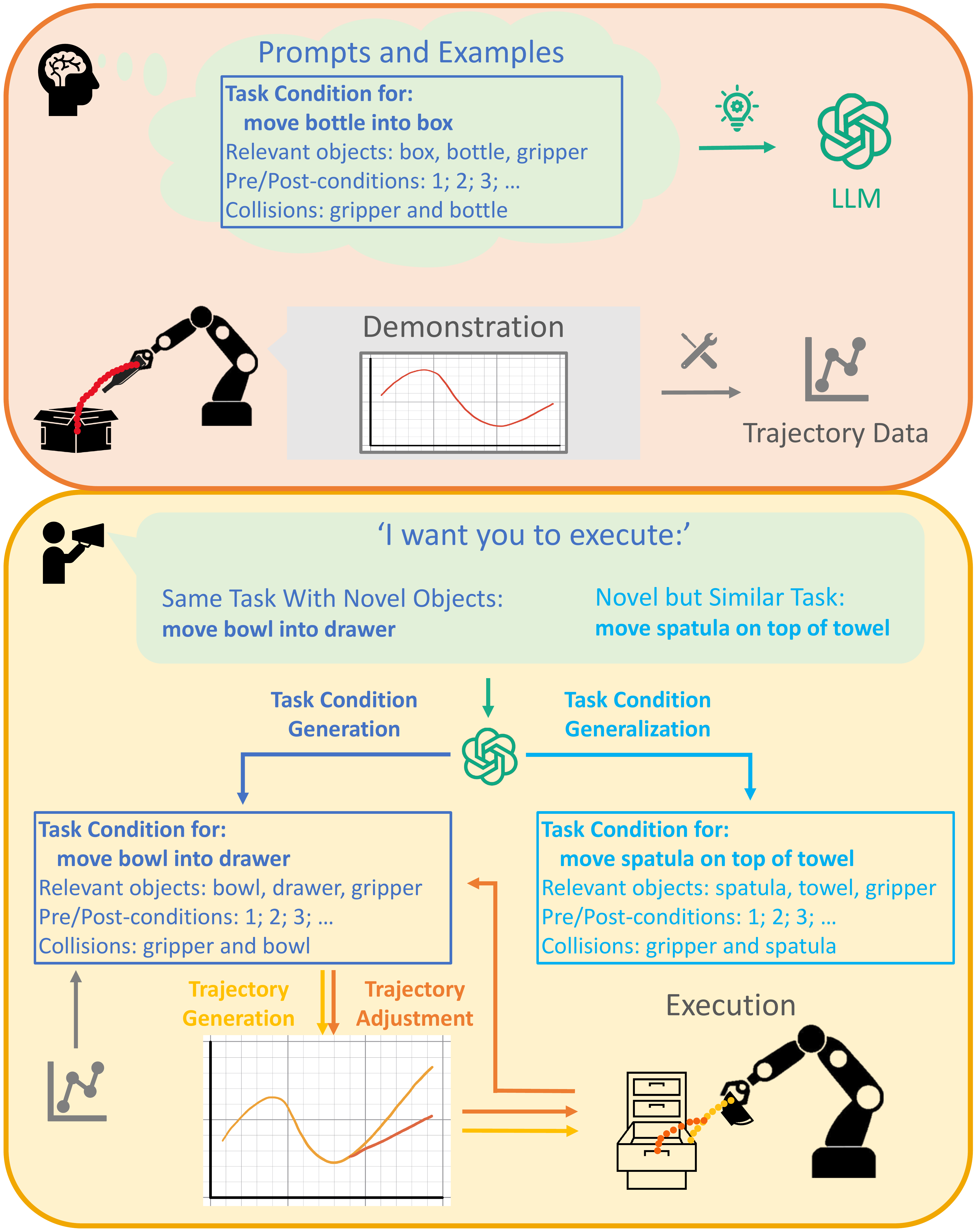}
    \label{fig: teaser figure}
    \vspace{-16pt}
    \caption{\textbf{Overall Framework.} 
    We leverage LLMs to generate and generalize primitive task conditions for both familiar tasks with novel objects and novel but related tasks.
    Subsequently, the high-level task conditions guide the generation and adjustment of low-level trajectories originally learned from demonstrations for long-horizon task execution.
    }
    \vspace{-18pt}
\end{figure}

While such environmental conditions can be obtained through data sampling and point cloud or image processing from human demonstrations, two key limitations arise. 
Firstly, acquiring such data and demonstrations can be costly; and secondly, it is challenging to generalize to novel scenarios and tasks without additional demonstrations. 
Recently, Large Language Models (LLMs) have demonstrated strong reasoning abilities with human commonsense and in-context learning capabilities. 
For example, \cite{liu2023reflect,singh2022progprompt,huang2022inner,huang2023grounded} showcase LLMs excel in failure explanation, task decomposition, and plan scoring for robot manipulation tasks.

Motivated by the above observations, this work leverages LLMs to create and generalize task conditions for both familiar tasks involving novel objects and new but related tasks.
This offers promise in two ways:
1) Task conditions can be derived not only from human demonstrations but also from the inherent commonsense knowledge within LLMs, and
2) The prior knowledge and in-context learning capabilities of LLMs help generalize to novel objects and unseen tasks.
Subsequently, the high-level task conditions guide the generation and adjustment of low-level Dynamic Movement Primitives (DMP) trajectories initially learned from demonstrations for long-horizon task execution.
These components are seamlessly integrated and rigorously evaluated in our proposed manipulation task suite simulator and real-world environments.
We summarize our \textbf{contributions} as follows:
\begin{itemize}
\setlength\itemsep{0.3em}
    \item An LLM module for generating and generalizing task conditions for both seen and unseen primitive tasks.
    \item A systematic long-horizon manipulation framework that leverages high-level task conditions to guide the generation of low-level action trajectory based on DMPs.
    \item A challenging manipulation task suite within Pybullet.
    \item Extensive experiments in both simulated and real-world settings to demonstrate the effectiveness of our pipeline.
\end{itemize}




\section{Related Work} \label{sec: related work}
\subsection{Robotic Skill Learning for Long Horizon Tasks}
A common approach to tackle long-horizon tasks is task-and-motion-planning~(TAMP) that decomposes the planning process of a long-horizon task into discrete symbolic states and continuous motion  generation~\cite{kaelbling2011hierarchical,migimatsu2020object,toussaint2015logic}. However, classical TAMP methods rely on manually specified symbolic rules, thereby requiring known physical states with high dimensional search space in complex tasks. 
Recent works~\cite{Nair2020Hierarchical,pertch2020long} integrate learning into the TAMP framework to speed up the search of feasible plans~\cite{chitnis2016guided,kim2019learning,wang2018active} or directly predict the action sequences from an initial image~\cite{driess2020deep}. 
%

Another potential solution for solving long-horizon tasks is hierarchical reinforcement/imitation  learning~\cite{barto2003recent,kulkarni2016hierarchical}. 
The options framework~\cite{sutton1999between} and FUN (FeUdal Networks)~\cite{vezhnevets2017feudal}, adopt modular architectures where agents learn both high- and low-level policies, with the high-level policy generating abstract sub-tasks and the low-level policy executing primitive actions.
However, designing an efficient hierarchical structure and representing reusable domain knowledge across tasks remains a challenging problem.
Relay Policy Learning~\cite{gupta2020relay} involves an imitation learning stage that produces goal-conditioned hierarchical policies, followed by a reinforcement learning phase that fine-tunes these policies to improve task performance.
In this work, we explore using large language models as task condition generators for long-horizon robot manipulation, which are generalizable to both novel objects and novel tasks.

\subsection{Large Language Models for Robotic Learning}
LLMs, such as GPT-3~\cite{Brown-NeurIPS-2020-Language}, PaLM~\cite{Chowdhery-arxiv-2022-PaLM,Anil-arxiv-2023-palm2},  Galactica~\cite{Taylor-arxiv-2022-Galactica}, and LLaMA~\cite{Touvron-arxiv-2023-LLaMA}, exhibit strong capacities to understand natural language and solve complex tasks.
For robotics, many research endeavors have focused on enabling robots and other agents to comprehend and follow natural language instructions~\cite{Luketina2019ASO,ding2023embodied}, often through the acquisition of language-conditioned policies~\cite{shao2020concept,stepputtis2020language, nair2022learning,mees2022calvin,mees2022matters,jang2022bc,shridhar2022perceiver,brohan2023rt1}. 
Additionally, researchers also explores connecting LLMs to robot commands~\cite{ahn2022saycan,huang2022inner,liang2022code,singh2022progprompt,brohan2023rt1,vemprala2023chatgpt,lin2023text2motion,bucker2022latte}, leveraging pre-trained language embeddings~\cite{hill2020human,shao2020concept,lynch2021grounding,nair2022learning,jang2022bc,jiang2022vima,huang2023voxposer,shridhar2022cliport,nair2022r3m,mu2023ec2} and pre-trained vision-language models~\cite{stone2023moo,mu2023embodiedgpt,brohan2023rt2} in robotic imitation learning.
The uniqueness of our work is to leverage GPT-3.5 to generate generalizable task conditions for long-horizon manipulation tasks.

\section{Technical Approach} \label{sec: tech approach}
Our work presents a framework that leverages pretrained large language models (LLMs) to generate and generalize task conditions of primitive tasks, which are subsequently used to guide the generation and refinement of the Dynamic Movement Primitives (DMP) trajectories for long-horizon task execution.
%
The framework is structured into four parts:
\textbf{1) task condition generation} (Sec.~\ref{sec: cond gen}), \textbf{2) generalization} (Sec.~\ref{sec: cond gene}), which use LLMs to acquire task conditions for both seen and unseen tasks; 
\textbf{3) trajectory learning} (Sec.~\ref{sec: DMP gen}) for generating DMP trajectories; and \textbf{4) trajectory generation \& adjustment} (Sec.~\ref{sec: DMP gen}), which incorporates information from task conditions and the environment.


\begin{figure*}[t]
    \centering
    \includegraphics[width=0.99\textwidth]{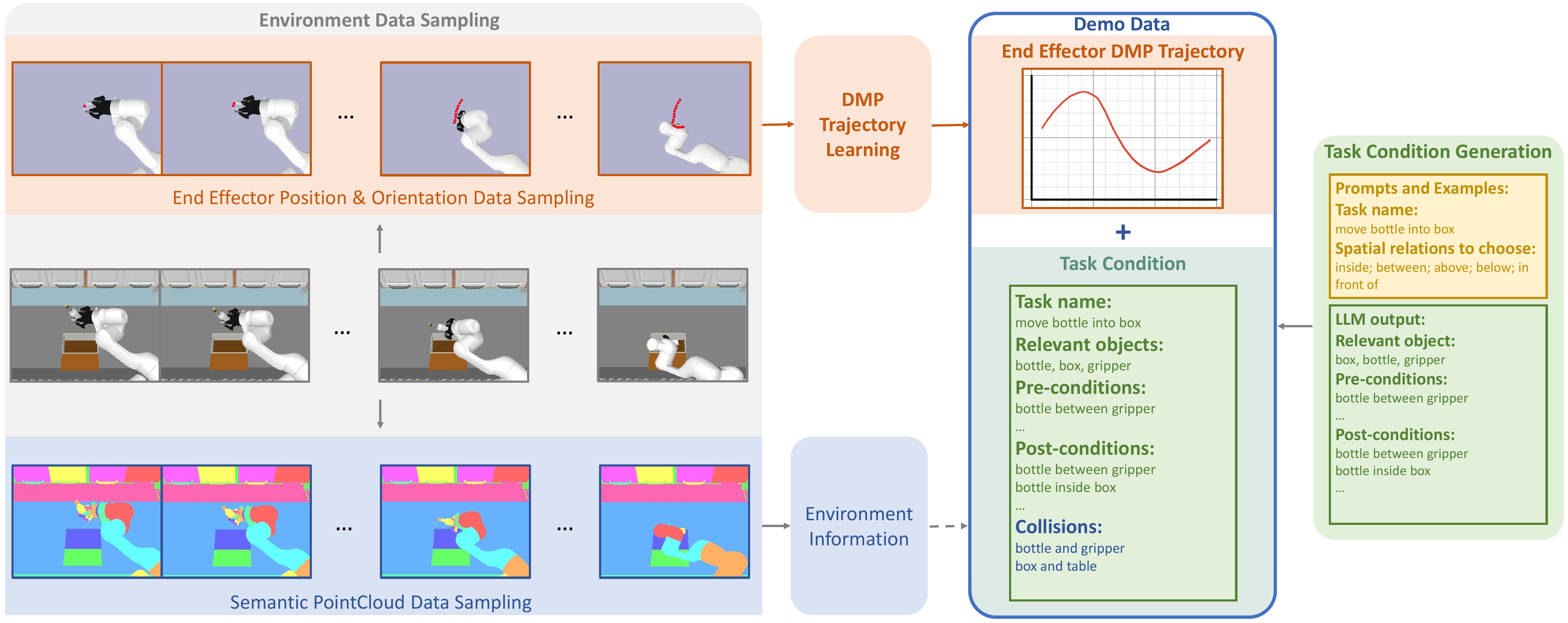}
    \vspace{-4pt}
    \caption{\textbf{Task Condition Generation and DMP Learning During Demonstration.} This figure shows the outline of our framework during demonstrations. Task condition is generated by LLM with prompts and examples (green part), or from environment information (blue part) as comparison. The trajectory is encoded by DMP (orange part). Task condition and trajectory make up the demo data.}
    \vspace{-10pt}
    \label{fig: LfD Structure}
\end{figure*}

\subsection{Definitions}  \label{sec: defs}

\subsubsection{Task Condition Definition} \label{sec: task cond definition}

In this work, the task condition for primitive tasks is focusing on the following aspects:
\begin{itemize}
\small
\setlength\itemsep{0.3em}
    \item \textbf{Task name:} a brief description of the primitive task.
    \item \textbf{Relevant objects:} potential objects relevant to the primitive task.
    \item \textbf{Pre-conditions:} environmental conditions that must be satisfied for the initiation of a primitive task, otherwise lead to task failure.
    \item \textbf{Post-conditions:} environmental conditions that mark the success of a primitive task.
    \item \textbf{Collisions:} permissible collisions during the task execution.
\end{itemize}

\begin{lstlisting}[
float=t,
language=Python,
floatplacement=htbp,
frame=single,
frameround=tftf,
belowskip=-3\baselineskip,
basicstyle=\ttfamily\notsotiny,
breakatwhitespace=false,
breaklines=true,
captionpos=b,
keepspaces=true,
showspaces=false,
showstringspaces=false,
showtabs=false,
label={lst: Task condition structure},
caption=Task condition structure.]
<@\textbf{Task name:} move bottle into box@>
<@\textbf{Relevant objects:} bottle, box, gripper@>
<@\textbf{Pre-conditions:}@>
<@  \textcolor{magenta}{gripper grasping bottle}@>
<@  \textcolor{magenta}{box open}@>
<@\textbf{Post-conditions:}@>
<@  \textcolor{orange}{bottle inside box}@>
<@  \textcolor{orange}{gripper inside box}@>
<@  \textcolor{magenta}{gripper grasping bottle}@>
<@  \textcolor{magenta}{box open}@>
<@\textbf{Collisions:} \textcolor{orange}{gripper and bottle}@>
\end{lstlisting}

A typical example of task condition is shown in Lst.~\ref{lst: Task condition structure}, the pre/post-conditions are divided into two categories:
\begin{itemize}
\small
\setlength\itemsep{0.3em}
    \item \textbf{Spatial relations:} shown in \textcolor{orange}{orange}, representing spatial relations between relevant objects. The spatial relations we will focus on include: \textit{inside, above, below, in front of} 
    \item \textbf{Object states:} shown in \textcolor{magenta}{magenta}, representing the state of the relevant objects. The object states we will focus on include: \textit{gripper: grasping/not grasping, and container: open/close}
\end{itemize}

\subsubsection{Environment Information Acquiring} \label{sec: env info}
We enable the framework with a certain perception ability. The raw data we are getting from the environment is semantic point clouds generated from RGBD cameras. The following methods are then used to get different environmental information:
\begin{itemize}
\small
\setlength\itemsep{0.3em}
    \item \textbf{Object center position/bounding box:} we get the ground truth object center position and bounding box directly from Pybullet if it is visible in the semantic point cloud. A Gaussian noise is then added to each coordinate of the center position and bounding box, in order to simulate error in the real scenario.
    \item \textbf{Spatial relations:} the four spatial relations between objects mentioned in Sec.~\ref{sec: task cond definition} is generated based on center positions and bounding boxes of the objects, as in~\cite{liu2023reflect}.
    \item \textbf{Object states:} the gripper state is judged through its spatial relations with other objects. If another is inside the gripper, we consider it to be grasped; the open or closed state of containers is judged by getting their moving joint position.
    \item \textbf{Collisions:} Collision between environment objects is detected using an implemented function in Pybullet. The position of the collision is generated at the same time.
\end{itemize}

\subsubsection{Trajectory and Manipulator Control} \label{sec: traj and mani control}
In our framework, we sample and generate manipulator trajectories using end-effector pose in Cartesian coordinates and Euler angles.
For manipulator control, we convert between Cartesian and joint space using Pybullet's built-in inverse kinematics. We employ position control, which, while straightforward, proves to be stable for the defined primitive tasks.
\subsection{Task Condition Generation with LLMs} \label{sec: cond gen}
\begin{lstlisting}[
float=t,
language=Python,
floatplacement=htbp,
frame=single,
frameround=tftf,
belowskip=-1\baselineskip,
basicstyle=\ttfamily\notsotiny,
breakatwhitespace=true,
breaklines=true,
captionpos=b,
keepspaces=true,
showspaces=false,
showstringspaces=false,
showtabs=false,
label={lst: Prompt for SR},
caption=Prompt for spatial relation \& collision generation.,
abovecaptionskip=0pt,
belowcaptionskip=0pt]
<@Imagine you are a spatial relation \& collision judgment machine@>
<@\textcolor{blue}{Here are the spatial relations you can choose from:}@>
  <@\textcolor{blue}{above, below, inside, in front of}@>
<@\textcolor{orange}{Here are some examples:}@>
<@\textcolor{orange}{{[}Task condition examples with only spatial relation \& collision{]}}@>
<@\textcolor{Lavender}{Remember to strictly follow the examples' format.}@>
<@\textcolor{teal}{Q: Task name: {[}Generation Task Name{]}}
\end{lstlisting}

\begin{lstlisting}[
float=t,
language=Python,
floatplacement=htbp,
frame=single,
frameround=tftf,
belowskip=-2\baselineskip,
basicstyle=\ttfamily\notsotiny,
breakatwhitespace=false,
breaklines=true,
captionpos=b,
keepspaces=true,
showspaces=false,
showstringspaces=false,
showtabs=false,
label={lst: Prompt for OS},
caption=Prompt for object state generation.,
abovecaptionskip=0pt,
belowcaptionskip=0pt]
<@Imagine you are an object state judgment machine@>
<@\textcolor{blue}{Here are the object states you can choose from:}@>
  <@\textcolor{blue}{gripper grasping/not grasping, container open/closed}@>
<@\textcolor{Mahogany}{Here are some examples:}@>
<@\textcolor{Mahogany}{{[}Examples of task condition with only object states{]}}@>
<@\textcolor{Lavender}{Remember to strictly follow the examples' format.}@>
<@\textcolor{teal}{Q: Task name: {[}Generation Task Name{]}}
\end{lstlisting}

In order to generate task conditions of the same task with novel objects using LLM, we must overcome the well-known shortcoming of it being random and inconsistent. We managed to achieve this goal by first dividing the task condition generation problem into two parts: spatial relations \& collisions generation problem, and object states generation problem. Two LLM chats are prompted differently to do these two parts separately. Also, the prompts we design emphasize the importance of remembering the examples given to it, and generating answers in the same format.

The prompt for spatial relations and collision generation is shown in Lst.~\ref{lst: Prompt for SR}. It has four components overall, including:
\begin{itemize}
\small
\setlength\itemsep{0.3em}
    \item an overall description of the job for LLM, indicated in black;
    \item spatial relations or object states to choose from, indicated in \textcolor{blue}{blue};
    \item examples in the form of Lst.~\ref{lst: Task condition structure} with only spatial relations included, indicated in \textcolor{orange}{orange};
    \item answer format indicated in \textcolor{Lavender}{pink}.
\end{itemize}

Finally, we present a task name to LLM for task condition generation indicated by \textcolor{teal}{teal}. To be mentioned, object names will be represented by letters (A, B, etc.) instead of specific object names. The prompt for object state generation remains consistent in structure and composition, except that changes in descriptions and examples are made for object state generation. Detailed prompt is shown in Lst.~\ref{lst: Prompt for OS}.

\subsection{Task Condition Generalization with LLMs} \label{sec: cond gene}

\begin{lstlisting}[
float=t,
language=bash,
floatplacement=htbp,
frame=single,
frameround=tftf,
belowskip=-2\baselineskip,
basicstyle=\ttfamily\notsotiny,
breakatwhitespace=true,
breaklines=true,
captionpos=b,
columns=flexible,
keepspaces=true,
tabsize=2,
showspaces=false,
showstringspaces=false,
showtabs=false,
label={lst: Prompt for SR Gene},
caption= Chain-of-thoughts prompt for condition generalization.,
abovecaptionskip=0pt,
belowcaptionskip=0pt]
<@Imagine you are a spatial relation \& collision judgment machine@>
<@\textcolor{Purple}{I will give you a task name describing a manipulator task, the}@>
<@\textcolor{Purple}{end effector of the manipulator is a gripper.}@>
<@\textcolor{Purple}{First, you should determine what are the relevant objects in }@>
<@\textcolor{Purple}{this task}.@>
<@\textcolor{Purple}{Then, you should present what spatial relations these objects }@>
<@\textcolor{Purple}{should have before (pre-conditions) and after (post-conditions)}@>
<@\textcolor{Purple}{the task is processed.}@>
<@\textcolor{blue}{Here are the spatial relations you can choose from:}@>
  <@\textcolor{blue}{above, below, inside, in front of}@>
<@\textcolor{Purple}{At last, you should present what collisions there might occur}@>
<@\textcolor{Purple}{during the completion of the task. You must only generate}@>
<@\textcolor{Purple}{collisions that include the relevant objects.}@>
<@\textcolor{orange}{Here are some examples:}@>
<@\textcolor{orange}{{[}Task condition examples with only spatial relation \& collision{]}}@>
<@\textcolor{Lavender}{Remember to strictly follow the examples' format}@>
<@\textcolor{Purple}{Do not generate any pre/post conditions except spatial relations}@>
<@\textcolor{violet}{I gave you to choose from.}@>
<@\textcolor{teal}{Q: Task name: {[}Generalization Task Name{]}}
\end{lstlisting}

Task condition generalization with LLM follows the prompting concept we mentioned in Sec.~\ref{sec: cond gen}, but should be able to generalize novel but similar task conditions. Therefore, changes in prompts are made in order to let the LLM loosen up a little in order to leverage its reasoning and learning ability from examples, meanwhile making sure the answer is in the same format as examples. We add the chain of thoughts prompting into the previous prompt in Lst.~\ref{lst: Prompt for SR}, telling the LLM how to reason step by step, indicated in \textcolor{Purple}{purple} in Lst.~\ref{lst: Prompt for SR Gene}. 
The prompt for object states is similar and not presented here due to space constraints.


\subsection{Trajectory Learning with DMP} \label{sec: DMP learning}
Dynamic Movement Primitives (DMP) is a trajectory imitation learning method with high nonlinear characteristics and real-time performance ability. Meanwhile, it is capable of generating similar shaping trajectories to different goal positions.
A large amount of work has been done based on the original formulation raised by~\cite{schaal2006dynamic, ijspeert2013dynamical}, and we choose to stick with the original discrete DMP, considering it is aimed to solve trajectory optimization problems in Cartesian space, and shows simplicity yet efficiency in our framework. 

The basic formula of the discrete DMP is described by Eq.~\ref{eq: basic fomula}. \(y\) is the current system status (e.g. position) and \(\dot{y}\), \(\ddot{y}\) being its first and second derivatives. \(g\) is the goal status. The first term on the right is a PD controller with \(\alpha_y\) and \(\beta_y\) representing the P parameter and D parameter and \(\tau\) controlling the speed of convergence.
\begin{equation}
    \tau^2\ddot{y}=\alpha_y\left(\beta_y\left(g-y\right)-\tau\dot{y}\right)+f
    \label{eq: basic fomula}
\end{equation}

The nonlinear term \(f\) is implemented by the normalized weighting of multiple nonlinear basis functions to control the process of convergence. The variable \(x\) in it satisfies a first-order system, making the nonlinear term time-independent. Eventually, it can be described by Eq.~\ref{eq: nonlinear term}. \(N\) and \(w_i\) are the number and weight of the basis functions \(\Psi_i\) respectively, which we choose to be the Gaussian basis function described by Eq.~\ref{eq: basis funtion}, \(\sigma\) and \(c\) are its width and center position.
\begin{equation}   \label{eq: nonlinear term}
\small
    f(x,g)=\frac{\sum_{i=1}^{N}\Psi_i\left(x\right)w_i}{\sum_{i=1}^{N}\Psi_i\left(x\right)}x(g-y_0),~~~\tau\dot{x}=-\alpha_xx
\end{equation}
\begin{equation}
\small
    \Psi\left(x\right)=exp\left(\frac{-1}{2\sigma^2\left(x-c\right)^2}\right)
    \label{eq: basis funtion}
\end{equation}

\begin{figure}[t]
    \includegraphics[width=0.99\linewidth]{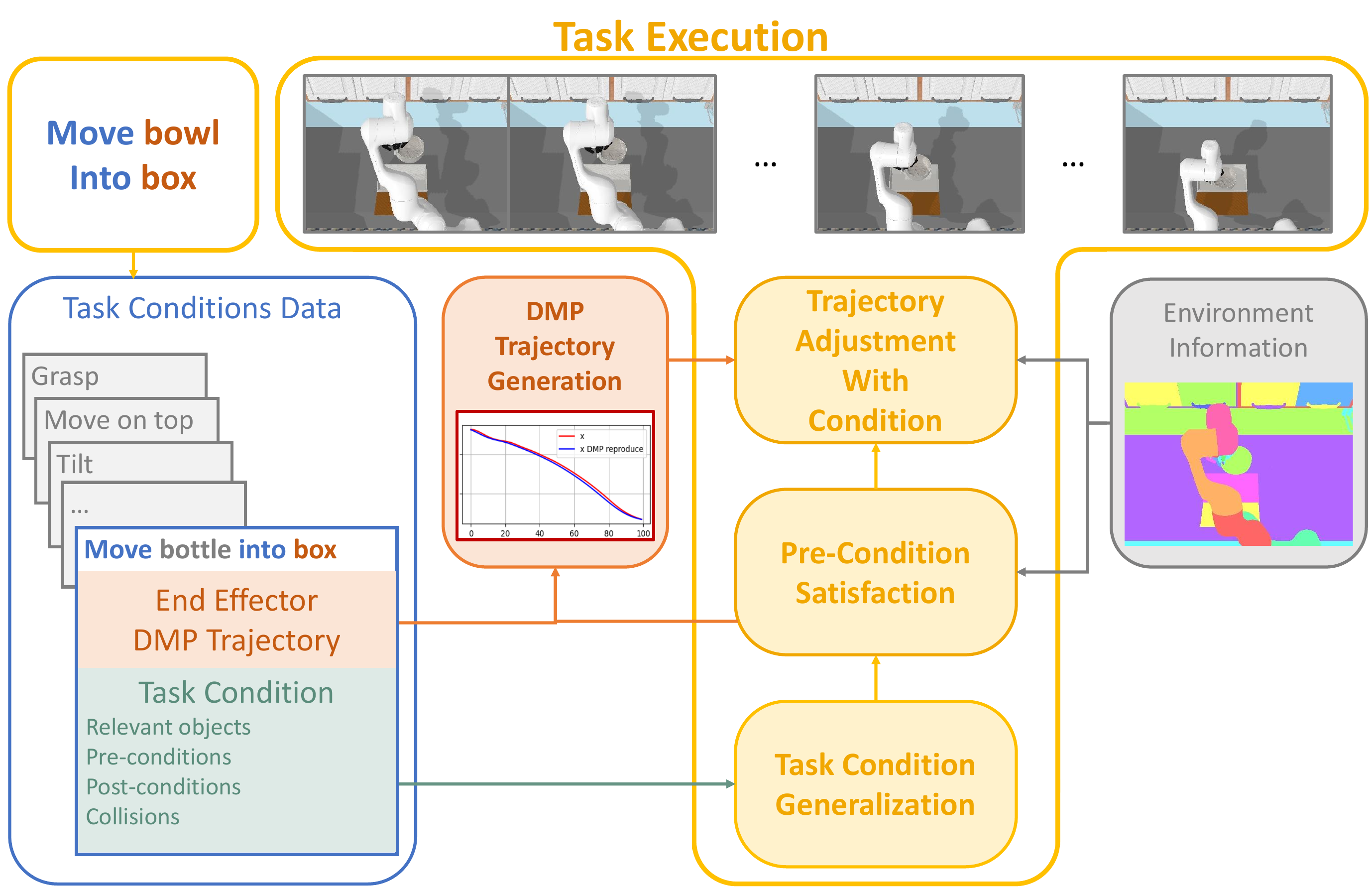}
    \vspace{-4pt}
    \caption{\textbf{Framework Flow Chat During Task Execution.} This figure starts with the top-left primitive task we wish to execute, and the task condition of it is generalized based on generated task conditions. Further, information in the task condition and from the environment is used to help DMP generate and adjust end-effector trajectory, achieving execution of the given primitive task.}
    \vspace{-8pt}
    \label{fig: execution structure}
\end{figure}

In our work, we sample the trajectory of the end effector \(y\), \(\dot{y}\) and \(\ddot{y}\) in each dimension mentioned in Sec.~\ref{sec: traj and mani control}, and the nonlinear term can be represented by rearranging Eq.~\ref{eq: basic fomula} and put \(f\) to the left side. Then, with hyperparameters in Eq.~\ref{eq: basic fomula} and Eq.~\ref{eq: basis funtion} set as: $N=100, \alpha_y=60, \beta_y=\ \frac{\alpha_y}{4}, \alpha_x=1, x_0=1$, we conducted the locally weighted regression method in ~\cite{ijspeert2013dynamical} to calculate rest of the variables, and mathematical expressions for calculation will not presented here. The trajectory encoded by DMP will be part of the demo data shown in Fig.~\ref{fig: LfD Structure}.


\begin{figure}[t]
\vspace{-6pt}
\begin{algorithm}[H]
\caption{\small Pre-Condition Satisfaction}
\label{alg: Task condition satisfy}
\footnotesize
\begin{algorithmic}[1]
\Require
\Statex \textit{T\_Cond} - \textbf{T}ask Condition
\Function {\colorbox[RGB]{224,224,224}{SatisfyPreCond}}{\textit{T\_Cond}}
    \State \textit{currentConds} $\gets$ \textproc{FromEnvPC}(\textit{})
    \State \textit{preConds} $\gets$ \textproc{FromCond}(\textit{T\_Cond})
    \For {\textit{preCond} \textbf{in} \textit{preConds}}
        \If {\textit{preCond} \textbf{not in} \textit{currentCond}}
            \State \textit{newT\_Name} $\gets$ \textproc{Cond2Task}(\textit{preCond})
            \State \textit{newT\_Cond} $\gets$ \textproc{GenCond}(\textit{newT\_Name})
            \State \textproc{\colorbox[RGB]{224,224,224}{SatisfyPreCond}}
            (\textit{newT\_Cond})
            \State \textproc{ControlWithCond}(\textit{newT\_Cond})
        \EndIf
        \If {\textit{RUNTIME} $>$ \textit{60s}}
            \State break
        \EndIf
    \EndFor
\EndFunction
\end{algorithmic}
\end{algorithm}
\vspace{-24pt}
\end{figure}
\begin{figure}[t]
\begin{algorithm}[H]
\caption{\small Trajectory Generation With Condition}
\label{alg: control_manipulator}
\footnotesize
\begin{algorithmic}[1]
\Require{\textit{T\_Cond} - \textbf{T}ask Condition}
\Function{ControlWithCond}{\textit{T\_Cond}}
\State \textit{taskColl, postConds} $\gets$ FromCond(\textit{T\_Cond})
\State \textit{targetPos} $\gets$ FromEnvPC()
\State \textit{goalPos} $\gets$ \textit{targetObjPos}
\State \textit{trajDMP}[:] $\gets$ GenDMP(\textit{goalPos})
\While {\textit{RUNTIME} $\leq$ \textit{60s} \textbf{or} end(\textit{trajDMP}) \textbf{or} postConds satisfied} 
    \State ControlManipulator(\textit{trajDMP}[+1])
    \State \textit{currentColl} $\gets$ FromEnvPC()
    \If {\textit{currentColl} not in \textit{taskColl}}
        \State ControlManipulator(\textit{trajDMP}[-10])
        \State \textit{collPos} $\gets$ FromEnvPC()
        \State \textit{goalPos} $\gets$ \textit{goalPos} + (\textit{goalPos} - \textit{collPos})
        \State \textit{trajDMP}[:-1] $\gets$ GenDMP(\textit{goalPos})
    \EndIf
\EndWhile
\EndFunction
\end{algorithmic}
\end{algorithm}
\vspace{-24pt}
\end{figure}

\subsection{Trajectory Generation \& Adjustment via Task Conditions} \label{sec: DMP gen}
The method to generate trajectories using DMP is straightforward with parameters learned in Sec.~\ref{sec: DMP learning}. With new starting position \(y_0\) and goal position \(g\) given, we can use them to generate the nonlinear term described by Eq.~\ref{eq: nonlinear term} as well as position at any point by running a numerical simulation of the second-order system described by Eq.~\ref{eq: basic fomula}. In our framework, this method will be guided by corresponding task conditions to help generate and adjust trajectory. Two main functions are used sequentially, which are shown as yellow blocks in Fig.~\ref{fig: execution structure}.

The first function is pre-condition satisfaction. It is conducted before reproducing a primitive task and is aimed to satisfy the pre-conditions of the task before execution. To achieve this, a recursive algorithm is used and its pseudo code is shown as Alg.~\ref{alg: Task condition satisfy}. Given the task condition, the algorithm gets the pre-conditions in it and compares them to the current conditions in the environment acquired from environment information mentioned in Sec.~\ref{sec: env info}. Then, the pre-conditions that are not currently satisfied will raise a primitive task aiming to satisfy it. The mapping between conditions and primitive tasks is done by humans in advance, for example, `gripper grasping bottle' not satisfied will lead to `grasp bottle' as a new primitive task, and its pre-condition will be generated and satisfied recursively.

The second function is trajectory adjustment with task conditions to prevent unintended collisions during execution. 
The pseudo-code is shown as Alg.~\ref{alg: control_manipulator}. First, we generate a trajectory given the target object center from environment information as goal position using DMP, primitive tasks without a target object, such as moving or folding, goal position is defined by the difference between starting and ending positions during demonstration. During the execution of the trajectory, we monitor collisions in the environment. If there is an unwanted collision not in task condition, we stop executing and go backward 10 trajectory points. Then, we update the goal position regarding where the collision happens and adjust the rest of the trajectory. 
The algorithm terminates when all post-conditions are met or if the execution time exceeds a predefined threshold.

\section{Simulation} \label{sec: sim}
To evaluate our framework, we design a challenging Robotic Manipulation Task Suite in Pybullet~\cite{coumans2021}. The environment consists of two 7 Dof robots Franka and Kinova with a kitchen scene including various interactive objects. It contains 10 diverse primitive tasks (37 if considering different objects) and 4 long-horizon tasks in simulation.

These primitive tasks cover a broad spectrum of robotic manipulation skills. Our simulation includes 1) rigid object manipulation such as grasping and moving, 2) articulated object manipulation such as open/close the drawer/box, 3) periodic manipulation such as tilting the mug, 4) soft object manipulation such as grasp the cloth, and 5) dual-arm manipulation tasks, \emph{e.g.}, folding the cloth. The detailed description of primitive tasks and long-horizon tasks is shown in Tab.~\ref{tab: Task descriptions}.
We visualize our introduced simulator in Fig.~\ref{fig:real_world} (left).

\begin{table}[t]
\vspace{3pt}
\caption{Task descriptions. Notations: PT -- primitive task, LHT -- long-horizon task, * -- primitive task with novel objects.}
\vspace{-4pt}
\centering
\scriptsize
\begin{tabular}{@{}p{0.075\linewidth}|p{0.22\linewidth}?p{0.075\linewidth}|p{0.43\linewidth}}
\toprule[1pt]
PT1 & Grasp Object  & PT6 & Fold Object\\
PT2 & Release Object & PT7 & Move Object A to position \\
PT3 & Open Object & PT8 & Move Object A On Top of Object B \\
PT4 & Close Object & PT9 & Move Object A into Object B\\
PT5 & Tilt Object & PT10 & Move Object A In Front of Object B \\
\midrule[0.7pt]
LHT1 & \multicolumn{3}{l}{Pick up a bottle, put the bottle into the box, close the box.}\\
\tabincell{l}{LHT2 \\~\\~} & \multicolumn{3}{l}{\tabincell{l}{To put the bowl into the bottom drawer of the cabinet, first close the \\ top drawer, then open the bottom drawer, put the bowl into the bottom \\ drawer and then close the drawer.}}\\
LHT3 & \multicolumn{3}{l}{Hold a mug up, place it on the table, and put toothpaste into the mug.}\\
LHT4 & \multicolumn{3}{l}{Grasp the spatula and place it onto the cloth, then fold the cloth.}\\
\midrule[0.7pt]
LHT1* & \multicolumn{3}{l}{Pick up a bowl, put the bowl into the box, and close the box.}\\
\tabincell{l}{LHT2* \\~\\~} & \multicolumn{3}{l}{\tabincell{l}{To put the bowl into the upper drawer of the cabinet, first close the \\ bottom drawer, then open the upper drawer, put the bowl into the upper \\ drawer and then close the drawer.}}\\
\bottomrule[1pt]
\end{tabular}
\vspace{-12pt}
\label{tab: Task descriptions}
\end{table}

\section{Experiments} \label{sec: exp}
\subsection{Task Condition Generation \& Generalization Experiment} \label{sec: cond gen exp}

First, we evaluate the ability of our framework to generate and generalize task conditions on all 10 primitive tasks. The LLM (GPT-3.5) is provided with condition examples. Comparison is made with task conditions generated from environments. A successfully generated task condition should contain accurate and enough information to guide the execution of the primitive task. The result is shown in Tab.~\ref{tab: condition generation success rate}.

\begin{table}[t]
\vspace{3pt}
\caption{Task condition generation and generalization.}
\vspace{-4pt}
\setlength{\tabcolsep}{8.1pt}
\renewcommand{\arraystretch}{0.97}
\label{tab: condition generation success rate}
\centering
\begin{tabular}{l|c c|c c}
\toprule
\multirow{2.5}{*}{\thead{PT Name}} & \multicolumn{2}{c|}{\thead{Generation}} & \multicolumn{2}{c}{\thead{Generalization}}\\ 
\cline{2-5}
& \thead{LLM} & \thead{FromEnv} & \thead{LLM} & \thead{FromEnv}\\
\midrule
\ Grasp & 100\% & 100\% & 24\% & - \\
\ Release & 100\% & 100\% & 36\% & - \\
\ Open & 90\% & \textbf{100\%} & 18\% & - \\
\ Close & 100\% & 100\% & 25\% & - \\
\ Tilt & 100\% & 100\% & 27\% & - \\
\ Fold & \textbf{100\%} & 95\% & 30\% & - \\
\ Move & 81\% & \textbf{100\%} & 40\% & - \\
\ MoveInTo & 90\% & \textbf{91.7\%} & 32\% & - \\
\ MoveOnTop & \textbf{90\%} & 86.7\% & 36\% & - \\
\ MoveInFront & 100\% & 100\% & 27\% & - \\
\bottomrule
\end{tabular}
\vspace{-4pt}
\end{table}

\begin{table}[t]
\caption{Primitive task execution success rate.}
\vspace{-4pt}
\label{tab: PT_success_rate}
\setlength{\tabcolsep}{10.2pt}
\centering
\begin{tabular}{c|c|c|c}
\toprule
\thead{PT Name} & \thead{Baseline} & \thead{w/o Cond} & \thead{w/\,\,\, Cond}\\\midrule
\ Grasp & 8.3\% & 87.5\% & \textbf{93.1}\% \\
\ Release & 100\% & 100\% & 100\% \\
\ Open & 6.7\% & \textbf{90\%} & \textbf{90\%} \\
\ Close & 31.7\% & \textbf{89.6\%} & \textbf{89.6\%} \\
\ Tilt  & \textbf{86.7\%} & 75\% & 85\% \\
\ Fold & 0\% & 80.2\% & \textbf{85\%} \\
\ Move & 16.65\% & 90.9\% & \textbf{94.1\%} \\
\ MoveInTo  & 68.8\% & 90.9\% & \textbf{94.4\%} \\
\ MoveOnTop & 45.6\% & 81.9\% & \textbf{91.1}\% \\
\ MoveInFront & 13.3\% & \textbf{100\%} & \textbf{100\%} \\
\bottomrule
\end{tabular}
\vspace{-12pt}
\end{table}

When we evaluate the result of the experiment, the high success rate in task generation shows our properly prompted LLM has consistent abilities to accurately determine relevant objects and generate correct task conditions in an expected format. It even outperformed the success rate of generation using environmental information in some cases. When it comes to task condition generalization, the LLM still has a chance to figure out proper task condition for every primitive task, with the highest success rate being 40\%. This shows our prompting method is feasible and gives the LLM task condition generalization ability. In contrast, another method clearly does not have the ability to generalize since no demonstration of the corresponding primitive task was given. 

\subsection{Primitive Task Execution Experiment} \label{sec: PT exp}

Secondly, we evaluate the ability of our framework to do imitation learning with DMP, and leverage task conditions to generate and adjust DMP trajectories. We run the execution of all 10 primitive tasks using three different methods:
\begin{itemize}
\small
\setlength\itemsep{0.3em}
    \item the image-based imitation learning baseline;
    \item with only DMP learning and generating trajectories; 
    \item with correct task conditions generating and adjusting trajectories. 
\end{itemize}

From Tab.~\ref{tab: PT_success_rate}, the image-based baseline achieves much worse performance in all but PT5 execution, with ups and downs from 0 to 100, since it highly depends on good quality images with certain angles. In contrast, trajectories generated by DMP alone show better robustness with a consistent over 80\% success rate in most primitive task execution. Then, when we compare methods [w/o task condition], we can see the success rate on every primitive task is better if not the same when using our framework's method to generate and adjust trajectories. Though the improvement it makes seems not exceptional, it will make an impact when the success rate is to be multiplied during a long-horizon task.

\subsection{Long-horizon Task Execution Experiment} \label{sec: PHT exp}
Finally, we combine previous experiments to evaluate the overall ability of our framework to generate, generalize task conditions, then use it to generate and adjust DMP trajectory.

\subsubsection{Experiment Setting}
In order to improve the integrity of our execution experiment, and to maximize evaluation of different modules in our framework, we conduct the experiments under three cases. Before experiments, demo data including task condition and encoded trajectory of each primitive task is gained from demonstrations of long-horizon tasks (LHT1 to LHT4). And this experiment will take the success rate in Exp.~\ref{sec: cond gen exp} into consideration as well. The setting of the three cases for execution is as follows:
\begin{itemize}
\small
\setlength\itemsep{0.3em}
    \item {\normalsize \textit{same as demonstration}}: LHT1 to LHT4 using the same objects as in the demonstrations, with all example task conditions provided to the LLM.
    \item {\normalsize \textit{generalize to novel objects}}: LHT1* and LHT2* with novel objects, with all example task conditions from LHT1 and LHT2.
    \item {\normalsize\textit{generalize to novel primitive tasks}}: 
    LHT1 to LHT4 with a randomly selected unseen (novel) primitive task, the example conditions excluding this primitive task are provided to LLM.
\end{itemize}

Similar to Exp.~\ref{sec: PT exp}, we run execution and evaluation under each case using four different methods:
\begin{itemize}
\small
\setlength\itemsep{0.3em}
    \item the image-based imitation learning baseline; 
    \item with only DMP learning and generating trajectories; 
    \item with task conditions from the environment to guide trajectories;
    \item with task conditions from LLM to guide trajectories.
\end{itemize}

\begin{table}[t]
\vspace{3pt}
\caption{Evaluation of long-horizon tasks. `-' for infeasible.}
\vspace{-4pt}
\label{tab: LHT_success_rate}
\setlength{\tabcolsep}{8pt}
\centering
\begin{tabular}{l|c|c|c|c}
\toprule
\thead{Task Name} & \thead{Baseline} & \thead{w/o Cond} & \thead{w/\,\,\, Cond \\ FromEnv} & \thead{w/\,\,\,\,\, Cond \\ from LLM} \\\midrule
\multicolumn{4}{l}{\emph{Same as demonstration}:} \\
LHT1 & 0.8\% & 50\% & 59.63\% & \textbf{60.75}\% \\
LHT2 & 0\% & 35\% & \textbf{43.72}\% & 32.47\% \\
LHT3 & 0\% & 20\% & \textbf{31.80}\% & 21.25\% \\
LHT4 & 0\% & 40\% & 45.30\% & \textbf{49.50}\% \\
\midrule
\multicolumn{4}{l}{\emph{Generalize to novel objects}:} \\
LHT1* & - & - & - & \textbf{48.6\%} \\
LHT2* & - & - & - & \textbf{28.8\%} \\
\midrule
\multicolumn{4}{l}{\emph{Generalize to novel primitive tasks}:} \\
LHT1 & - & - & - & \textbf{19.5\%} \\
LHT2 & - & - & - & \textbf{10.1\%} \\
LHT3 & - & - & - & \textbf{7.7\%} \\
LHT4 & - & - & - & \textbf{15.5\%} \\
\bottomrule
\end{tabular}
\vspace{-12pt}
\end{table}

\subsubsection{Experiment Results}
We first evaluate the result under \textit{same as demonstration} case in Tab.~\ref{tab: LHT_success_rate}. The performances of the baseline and with only DMP meet our expectations regarding Exp.~\ref{sec: PT exp}, since the success rate of a long-horizon task is approximately the product of that the primitive tasks have. The baseline performs even worse due to more inconsistent images during long-horizon tasks. As for methods [w/o task condition], the improvement in each primitive task adds up, leading to around 10\% better with task condition generating and adjusting the trajectories, even considering the chance for failure in task condition generating. The difference between using environment information and LLM to generate task conditions depends on the result in Exp.~\ref{sec: cond gen exp}, with LLM still showing great competitiveness, outperforming in two long-horizon executions. 

When it comes to other two cases, any slight backwardness of our framework before does not seem so important, as it is the only feasible method to do generalization to novel objects and primitive tasks. Only an average 8\% decrease is noticed when generalizing the same task to novel objects. While for novel primitive tasks, because of the success rate shown in Exp.~\ref{sec: cond gen exp} experiencing a massive drop, there is an average decrease of 27.5\%, leaving only around 13\% success rate. But considering its difficulty, this is already a breakthrough compared to our other methods.

\subsubsection{Advantage}
We highlight an interesting advantage of our framework in executing long-horizon tasks. It has the ability to generate the needed trajectory on its own if an over far-sighted primitive task is given. For example, if we ask the manipulator to move a bottle before giving the instruction on grasping it, our framework can generate a grasping trajectory autonomously. Not until grasping is successfully finished will the manipulator begin moving it. This shows a certain intelligence our framework has and from certain aspects, can mean a higher success rate in long-horizon task execution.

\subsection{Real-world Experiments}

\begin{figure}[t]
\vspace{4pt}
    \includegraphics[width=0.49\linewidth]{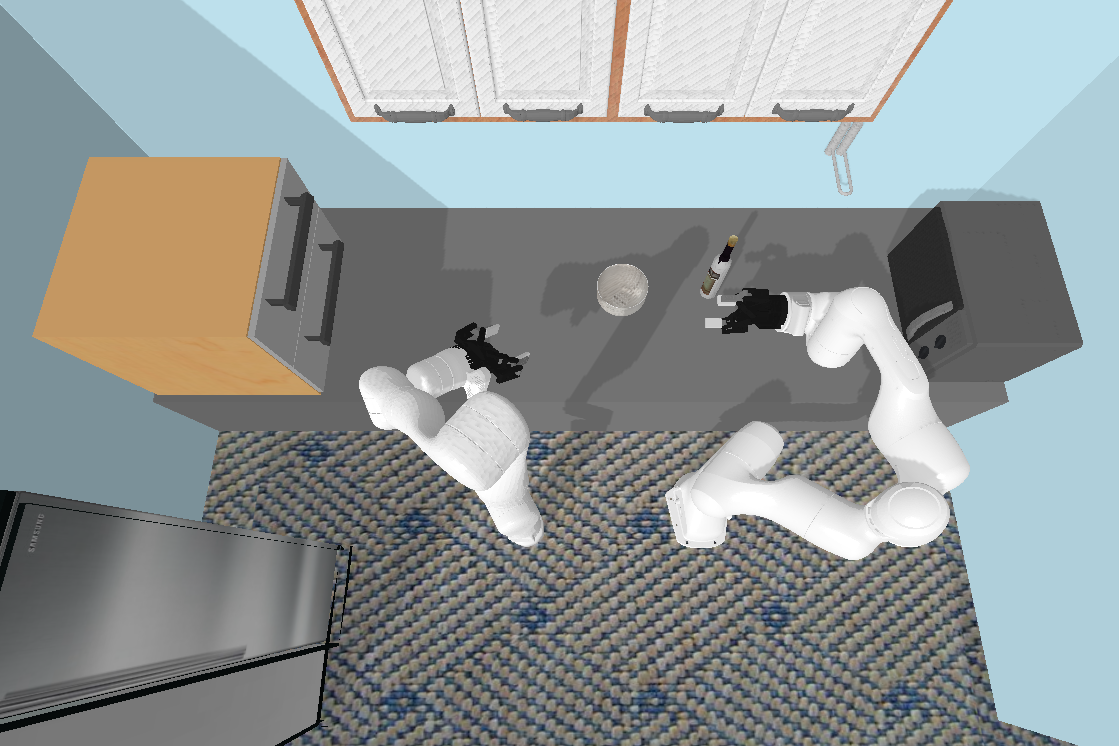}
    \includegraphics[width=0.49\linewidth]{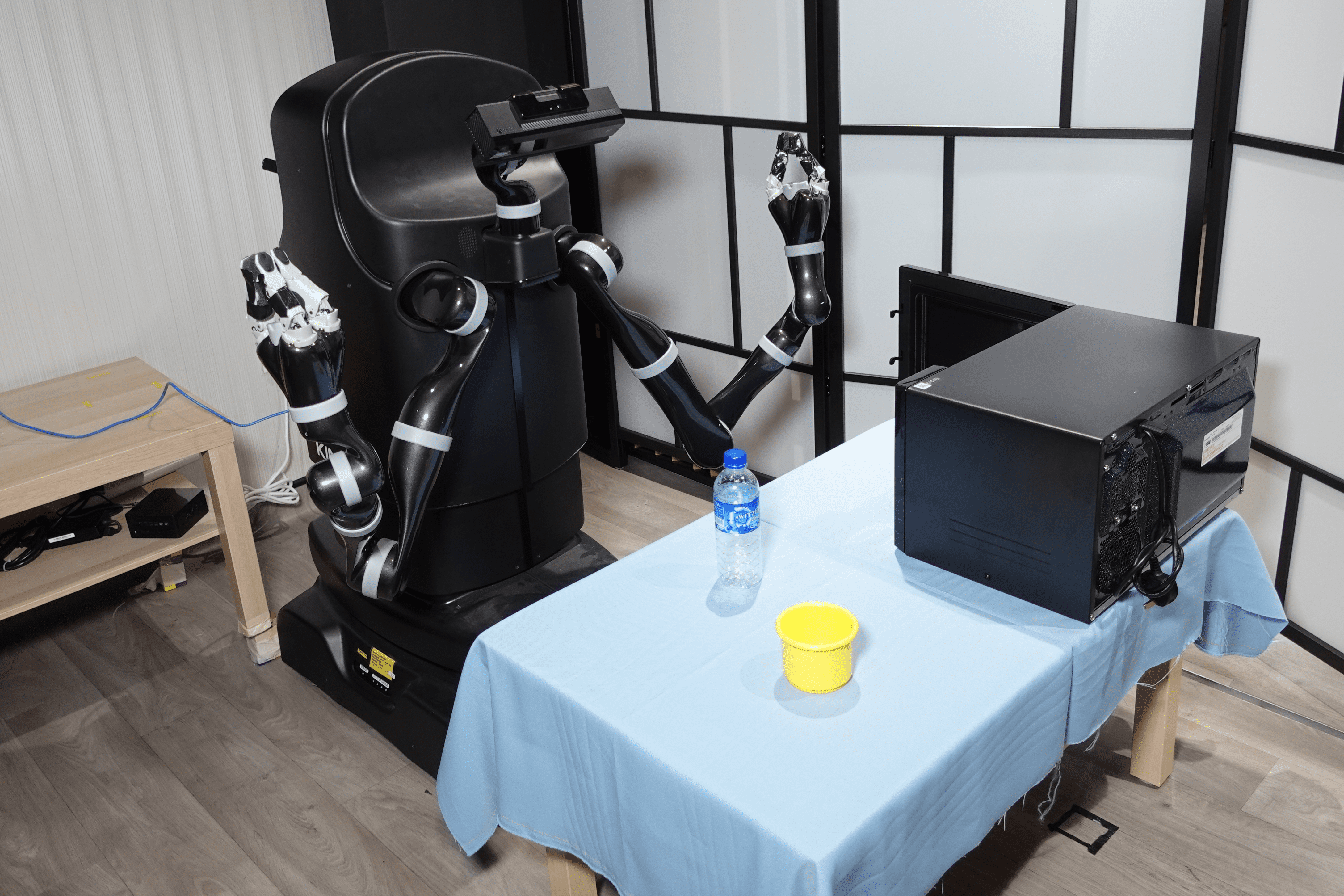}
    \vspace{-4pt}
    \caption{Our simulator vs. real-world experimental setup.}
    \label{fig:real_world}
    \vspace{-14pt}
\end{figure}

To demonstrate the practicality of our to perform long-horizon tasks in real scenarios, we set up an environment shown in Fig.~\ref{fig:real_world} (right). It shows objects such as a bowl, a bottle that the manipulator can grasp, and a microwave that the manipulator can open and close.
The environment features a dual-arm MOVO~\cite{movo} robot which has two 7 DoF manipulators and a Kinect RGB-D camera overhead. We use Segment Anything~\cite{kirillov2023segment} to obtain segmented point clouds of the surroundings and objects. Position control is performed as we use rangeIK~\cite{wang2023rangedik} for solving inverse kinematics. 
Due to space constraints, we present our real-world experiments in the supplemental video and the project website.

\section{Conclusion} \label{sec: conclusion}

This work explores the potential of LLMs on primitive task condition generalization for generalizable long-horizon manipulations with novel objects and unseen tasks.
The generated conditions are then utilized to steer the generation of low-level manipulation trajectories using DMP.
A robotic manipulation task suite based on Pybullet is also introduced.
We conduct experiments in both our simulation and real-world scenarios, demonstrating the effectiveness of the proposed framework for long-horizon manipulation tasks and its ability to generalize to tasks involving novel objects or unseen scenarios.
While our framework shows promise, there is room for improvement and a more versatile trajectory generator could complement our framework.

{\small
\bibliographystyle{IEEEtran}
\bibliography{references}

\begin{thebibliography}{10}
\providecommand{\url}[1]{#1}
\csname url@rmstyle\endcsname
\providecommand{\newblock}{\relax}
\providecommand{\bibinfo}[2]{#2}
\providecommand\BIBentrySTDinterwordspacing{\spaceskip=0pt\relax}
\providecommand\BIBentryALTinterwordstretchfactor{4}
\providecommand\BIBentryALTinterwordspacing{\spaceskip=\fontdimen2\font plus
\BIBentryALTinterwordstretchfactor\fontdimen3\font minus \fontdimen4\font\relax}
\providecommand\BIBforeignlanguage[2]{{%
\expandafter\ifx\csname l@#1\endcsname\relax
\typeout{** WARNING: IEEEtran.bst: No hyphenation pattern has been}%
\typeout{** loaded for the language `#1'. Using the pattern for}%
\typeout{** the default language instead.}%
\else
\language=\csname l@#1\endcsname
\fi
#2}}

\bibitem{kaelbling2011hierarchical}
L.~P. Kaelbling and T.~Lozano-P{\'e}rez, ``Hierarchical task and motion planning in the now,'' in \emph{2011 IEEE International Conference on Robotics and Automation}.\hskip 1em plus 0.5em minus 0.4em\relax IEEE, 2011, pp. 1470--1477.

\bibitem{migimatsu2020object}
T.~Migimatsu and J.~Bohg, ``Object-centric task and motion planning in dynamic environments,'' \emph{IEEE Robotics and Automation Letters}, vol.~5, no.~2, pp. 844--851, 2020.

\bibitem{toussaint2015logic}
M.~Toussaint, ``Logic-geometric programming: an optimization-based approach to combined task and motion planning,'' in \emph{Proceedings of the 24th International Conference on Artificial Intelligence}, 2015, pp. 1930--1936.

\bibitem{barto2003recent}
A.~G. Barto and S.~Mahadevan, ``Recent advances in hierarchical reinforcement learning,'' \emph{Discrete event dynamic systems}, vol.~13, no.~1, pp. 41--77, 2003.

\bibitem{kulkarni2016hierarchical}
T.~D. Kulkarni, K.~R. Narasimhan, A.~Saeedi, and J.~B. Tenenbaum, ``Hierarchical deep reinforcement learning: Integrating temporal abstraction and intrinsic motivation,'' \emph{Advances in Neural Information Processing Systems}, pp. 3682--3690, 2016.

\bibitem{liu2023reflect}
Z.~Liu, A.~Bahety, and S.~Song, ``Reflect: Summarizing robot experiences for failure explanation and correction,'' \emph{arXiv preprint arXiv:2306.15724}, 2023.

\bibitem{singh2022progprompt}
I.~Singh, V.~Blukis, A.~Mousavian, A.~Goyal, D.~Xu, J.~Tremblay, D.~Fox, J.~Thomason, and A.~Garg, ``Progprompt: Generating situated robot task plans using large language models,'' \emph{arXiv preprint arXiv:2209.11302}, 2022.

\bibitem{huang2022inner}
W.~Huang, F.~Xia, T.~Xiao, H.~Chan, J.~Liang, P.~Florence, A.~Zeng, J.~Tompson, I.~Mordatch, Y.~Chebotar, \emph{et~al.}, ``Inner monologue: Embodied reasoning through planning with language models,'' \emph{arXiv preprint arXiv:2207.05608}, 2022.

\bibitem{huang2023grounded}
W.~Huang, F.~Xia, D.~Shah, D.~Driess, A.~Zeng, Y.~Lu, P.~Florence, I.~Mordatch, S.~Levine, K.~Hausman, \emph{et~al.}, ``Grounded decoding: Guiding text generation with grounded models for robot control,'' \emph{arXiv preprint arXiv:2303.00855}, 2023.

\bibitem{Nair2020Hierarchical}
\BIBentryALTinterwordspacing
S.~Nair and C.~Finn, ``Hierarchical foresight: Self-supervised learning of long-horizon tasks via visual subgoal generation,'' in \emph{International Conference on Learning Representations}, 2020. [Online]. Available: \url{https://openreview.net/forum?id=H1gzR2VKDH}
\BIBentrySTDinterwordspacing

\bibitem{pertch2020long}
K.~Pertsch, O.~Rybkin, F.~Ebert, C.~Finn, D.~Jayaraman, and S.~Levine, ``Long-horizon visual planning with goal-conditioned hierarchical predictors,'' 2020.

\bibitem{chitnis2016guided}
R.~Chitnis, D.~Hadfield-Menell, A.~Gupta, S.~Srivastava, E.~Groshev, C.~Lin, and P.~Abbeel, ``Guided search for task and motion plans using learned heuristics,'' in \emph{2016 IEEE International Conference on Robotics and Automation (ICRA)}.\hskip 1em plus 0.5em minus 0.4em\relax IEEE, 2016, pp. 447--454.

\bibitem{kim2019learning}
B.~Kim, Z.~Wang, L.~P. Kaelbling, and T.~Lozano-P{\'e}rez, ``Learning to guide task and motion planning using score-space representation,'' \emph{The International Journal of Robotics Research}, vol.~38, no.~7, pp. 793--812, 2019.

\bibitem{wang2018active}
Z.~Wang, C.~R. Garrett, L.~P. Kaelbling, and T.~Lozano-P{\'e}rez, ``Active model learning and diverse action sampling for task and motion planning,'' in \emph{2018 IEEE/RSJ International Conference on Intelligent Robots and Systems (IROS)}.\hskip 1em plus 0.5em minus 0.4em\relax IEEE, 2018, pp. 4107--4114.

\bibitem{driess2020deep}
D.~Driess, J.-S. Ha, and M.~Toussaint, ``Deep visual reasoning: Learning to predict action sequences for task and motion planning from an initial scene image,'' in \emph{Robotics: Science and Systems 2020 (RSS 2020)}.\hskip 1em plus 0.5em minus 0.4em\relax RSS Foundation, 2020.

\bibitem{sutton1999between}
R.~S. Sutton, D.~Precup, and S.~Singh, ``Between mdps and semi-mdps: A framework for temporal abstraction in reinforcement learning,'' \emph{Artificial intelligence}, vol. 112, no. 1-2, pp. 181--211, 1999.

\bibitem{vezhnevets2017feudal}
A.~S. Vezhnevets, S.~Osindero, T.~Schaul, N.~Heess, M.~Jaderberg, D.~Silver, and K.~Kavukcuoglu, ``Feudal networks for hierarchical reinforcement learning,'' in \emph{International Conference on Machine Learning}.\hskip 1em plus 0.5em minus 0.4em\relax PMLR, 2017, pp. 3540--3549.

\bibitem{gupta2020relay}
A.~Gupta, V.~Kumar, C.~Lynch, S.~Levine, and K.~Hausman, ``Relay policy learning: Solving long-horizon tasks via imitation and reinforcement learning,'' in \emph{Conference on Robot Learning}.\hskip 1em plus 0.5em minus 0.4em\relax PMLR, 2020, pp. 1025--1037.

\bibitem{Brown-NeurIPS-2020-Language}
T.~B. Brown, B.~Mann, N.~Ryder, M.~Subbiah, J.~Kaplan, P.~Dhariwal, A.~Neelakantan, P.~Shyam, G.~Sastry, A.~Askell, S.~Agarwal, A.~Herbert{-}Voss, G.~Krueger, T.~Henighan, R.~Child, A.~Ramesh, D.~M. Ziegler, J.~Wu, C.~Winter, C.~Hesse, M.~Chen, E.~Sigler, M.~Litwin, S.~Gray, B.~Chess, J.~Clark, C.~Berner, S.~McCandlish, A.~Radford, I.~Sutskever, and D.~Amodei, ``Language models are few-shot learners,'' in \emph{Advances in Neural Information Processing Systems 33: Annual Conference on Neural Information Processing Systems 2020, NeurIPS 2020, December 6-12, 2020, virtual}, H.~Larochelle, M.~Ranzato, R.~Hadsell, M.~Balcan, and H.~Lin, Eds., 2020.

\bibitem{Chowdhery-arxiv-2022-PaLM}
A.~Chowdhery, S.~Narang, J.~Devlin, M.~Bosma, G.~Mishra, A.~Roberts, P.~Barham, H.~W. Chung, C.~Sutton, S.~Gehrmann, P.~Schuh, K.~Shi, S.~Tsvyashchenko, J.~Maynez, A.~Rao, P.~Barnes, Y.~Tay, N.~Shazeer, V.~Prabhakaran, E.~Reif, N.~Du, B.~Hutchinson, R.~Pope, J.~Bradbury, J.~Austin, M.~Isard, G.~Gur{-}Ari, P.~Yin, T.~Duke, A.~Levskaya, S.~Ghemawat, S.~Dev, H.~Michalewski, X.~Garcia, V.~Misra, K.~Robinson, L.~Fedus, D.~Zhou, D.~Ippolito, D.~Luan, H.~Lim, B.~Zoph, A.~Spiridonov, R.~Sepassi, D.~Dohan, S.~Agrawal, M.~Omernick, A.~M. Dai, T.~S. Pillai, M.~Pellat, A.~Lewkowycz, E.~Moreira, R.~Child, O.~Polozov, K.~Lee, Z.~Zhou, X.~Wang, B.~Saeta, M.~Diaz, O.~Firat, M.~Catasta, J.~Wei, K.~Meier{-}Hellstern, D.~Eck, J.~Dean, S.~Petrov, and N.~Fiedel, ``Palm: Scaling language modeling with pathways,'' \emph{CoRR}, vol. abs/2204.02311, 2022.

\bibitem{Anil-arxiv-2023-palm2}
R.~Anil, A.~M. Dai, O.~Firat, M.~Johnson, D.~Lepikhin, A.~Passos, S.~Shakeri, E.~Taropa, P.~Bailey, Z.~Chen, \emph{et~al.}, ``Palm 2 technical report,'' \emph{arXiv preprint arXiv:2305.10403}, 2023.

\bibitem{Taylor-arxiv-2022-Galactica}
R.~Taylor, M.~Kardas, G.~Cucurull, T.~Scialom, A.~Hartshorn, E.~Saravia, A.~Poulton, V.~Kerkez, and R.~Stojnic, ``Galactica: {A} large language model for science,'' \emph{CoRR}, vol. abs/2211.09085, 2022.

\bibitem{Touvron-arxiv-2023-LLaMA}
H.~Touvron, T.~Lavril, G.~Izacard, X.~Martinet, M.~Lachaux, T.~Lacroix, B.~Rozi{\`{e}}re, N.~Goyal, E.~Hambro, F.~Azhar, A.~Rodriguez, A.~Joulin, E.~Grave, and G.~Lample, ``Llama: Open and efficient foundation language models,'' \emph{CoRR}, 2023.

\bibitem{Luketina2019ASO}
J.~Luketina, N.~Nardelli, G.~Farquhar, J.~N. Foerster, J.~Andreas, E.~Grefenstette, S.~Whiteson, and T.~Rockt{\"a}schel, ``A survey of reinforcement learning informed by natural language,'' in \emph{IJCAI}, 2019.

\bibitem{ding2023embodied}
M.~Ding, Y.~Xu, Z.~Chen, D.~D. Cox, P.~Luo, J.~B. Tenenbaum, and C.~Gan, ``Embodied concept learner: Self-supervised learning of concepts and mapping through instruction following,'' in \emph{Conference on Robot Learning}.\hskip 1em plus 0.5em minus 0.4em\relax PMLR, 2023, pp. 1743--1754.

\bibitem{shao2020concept}
L.~Shao, T.~Migimatsu, Q.~Zhang, K.~Yang, and J.~Bohg, ``Concept2{R}obot: Learning manipulation concepts from instructions and human demonstrations,'' in \emph{Proceedings of Robotics: Science and Systems (RSS)}, 2020.

\bibitem{stepputtis2020language}
S.~Stepputtis, J.~Campbell, M.~Phielipp, S.~Lee, C.~Baral, and H.~Ben~Amor, ``Language-conditioned imitation learning for robot manipulation tasks,'' \emph{Advances in Neural Information Processing Systems}, vol.~33, pp. 13\,139--13\,150, 2020.

\bibitem{nair2022learning}
S.~Nair, E.~Mitchell, K.~Chen, S.~Savarese, C.~Finn, \emph{et~al.}, ``Learning language-conditioned robot behavior from offline data and crowd-sourced annotation,'' in \emph{Conference on Robot Learning}.\hskip 1em plus 0.5em minus 0.4em\relax PMLR, 2022, pp. 1303--1315.

\bibitem{mees2022calvin}
O.~Mees, L.~Hermann, E.~Rosete-Beas, and W.~Burgard, ``{CALVIN}: A benchmark for language-conditioned policy learning for long-horizon robot manipulation tasks,'' \emph{IEEE Robotics and Automation Letters}, 2022.

\bibitem{mees2022matters}
O.~Mees, L.~Hermann, and W.~Burgard, ``What matters in language conditioned robotic imitation learning over unstructured data,'' \emph{IEEE Robotics and Automation Letters}, vol.~7, no.~4, pp. 11\,205--11\,212, 2022.

\bibitem{jang2022bc}
E.~Jang, A.~Irpan, M.~Khansari, D.~Kappler, F.~Ebert, C.~Lynch, S.~Levine, and C.~Finn, ``{BC}-{Z}: Zero-shot task generalization with robotic imitation learning,'' in \emph{Conference on Robot Learning (CoRL)}, 2021, pp. 991--1002.

\bibitem{shridhar2022perceiver}
M.~Shridhar, L.~Manuelli, and D.~Fox, ``Perceiver-actor: A multi-task transformer for robotic manipulation,'' \emph{Conference on Robot Learning (CoRL)}, 2022.

\bibitem{brohan2023rt1}
A.~Brohan, N.~Brown, J.~Carbajal, Y.~Chebotar, J.~Dabis, C.~Finn, K.~Gopalakrishnan, K.~Hausman, A.~Herzog, J.~Hsu, \emph{et~al.}, ``{RT}-1: Robotics transformer for real-world control at scale,'' \emph{Robotics: Science and Systems (RSS)}, 2023.

\bibitem{ahn2022saycan}
M.~Ahn, A.~Brohan, N.~Brown, Y.~Chebotar, O.~Cortes, B.~David, C.~Finn, K.~Gopalakrishnan, K.~Hausman, A.~Herzog, \emph{et~al.}, ``Do as i can, not as i say: Grounding language in robotic affordances,'' \emph{Conference on Robot Learning (CoRL)}, 2022.

\bibitem{liang2022code}
J.~Liang, W.~Huang, F.~Xia, P.~Xu, K.~Hausman, B.~Ichter, P.~Florence, and A.~Zeng, ``Code as policies: Language model programs for embodied control,'' \emph{arXiv preprint arXiv:2209.07753}, 2022.

\bibitem{vemprala2023chatgpt}
S.~Vemprala, R.~Bonatti, A.~Bucker, and A.~Kapoor, ``Chat{GPT} for robotics: Design principles and model abilities,'' \emph{Microsoft Auton. Syst. Robot. Res}, vol.~2, p.~20, 2023.

\bibitem{lin2023text2motion}
K.~Lin, C.~Agia, T.~Migimatsu, M.~Pavone, and J.~Bohg, ``Text2motion: From natural language instructions to feasible plans,'' \emph{arXiv preprint arXiv:2303.12153}, 2023.

\bibitem{bucker2022latte}
A.~Bucker, L.~Figueredo, S.~Haddadin, A.~Kapoor, S.~Ma, and R.~Bonatti, ``Latte: Language trajectory transformer,'' \emph{arXiv preprint arXiv:2208.02918}, 2022.

\bibitem{hill2020human}
F.~Hill, S.~Mokra, N.~Wong, and T.~Harley, ``Human instruction-following with deep reinforcement learning via transfer-learning from text,'' \emph{arXiv preprint arXiv:2005.09382}, 2020.

\bibitem{lynch2021grounding}
C.~Lynch and P.~Sermanet, ``Grounding language in play,'' \emph{Robotics: Science and Systems (RSS)}, 2021.

\bibitem{jiang2022vima}
Y.~Jiang, A.~Gupta, Z.~Zhang, G.~Wang, Y.~Dou, Y.~Chen, L.~Fei-Fei, A.~Anandkumar, Y.~Zhu, and L.~Fan, ``{VIMA}: General robot manipulation with multimodal prompts,'' \emph{International Conference on Machine Learning (ICML)}, 2023.

\bibitem{huang2023voxposer}
W.~Huang, C.~Wang, R.~Zhang, Y.~Li, J.~Wu, and L.~Fei-Fei, ``Voxposer: Composable 3d value maps for robotic manipulation with language models,'' \emph{arXiv preprint arXiv:2307.05973}, 2023.

\bibitem{shridhar2022cliport}
M.~Shridhar, L.~Manuelli, and D.~Fox, ``Cliport: What and where pathways for robotic manipulation,'' in \emph{Conference on Robot Learning}.\hskip 1em plus 0.5em minus 0.4em\relax PMLR, 2022, pp. 894--906.

\bibitem{nair2022r3m}
S.~Nair, A.~Rajeswaran, V.~Kumar, C.~Finn, and A.~Gupta, ``R3m: A universal visual representation for robot manipulation,'' \emph{arXiv preprint arXiv:2203.12601}, 2022.

\bibitem{mu2023ec2}
Y.~Mu, S.~Yao, M.~Ding, P.~Luo, and C.~Gan, ``Ec2: Emergent communication for embodied control,'' in \emph{Proceedings of the IEEE/CVF Conference on Computer Vision and Pattern Recognition}, 2023, pp. 6704--6714.

\bibitem{stone2023moo}
A.~Stone, T.~Xiao, Y.~Lu, K.~Gopalakrishnan, K.-H. Lee, Q.~Vuong, P.~Wohlhart, B.~Zitkovich, F.~Xia, C.~Finn, \emph{et~al.}, ``Open-world object manipulation using pre-trained vision-language models,'' \emph{arXiv preprint arXiv:2303.00905}, 2023.

\bibitem{mu2023embodiedgpt}
Y.~Mu, Q.~Zhang, M.~Hu, W.~Wang, M.~Ding, J.~Jin, B.~Wang, J.~Dai, Y.~Qiao, and P.~Luo, ``Embodiedgpt: Vision-language pre-training via embodied chain of thought,'' \emph{arXiv preprint arXiv:2305.15021}, 2023.

\bibitem{brohan2023rt2}
A.~Brohan, N.~Brown, J.~Carbajal, Y.~Chebotar, X.~Chen, K.~Choromanski, T.~Ding, D.~Driess, A.~Dubey, C.~Finn, \emph{et~al.}, ``{RT}-2: Vision-language-action models transfer web knowledge to robotic control,'' \emph{arXiv preprint arXiv:2307.15818}, 2023.

\bibitem{schaal2006dynamic}
S.~Schaal, ``Dynamic movement primitives-a framework for motor control in humans and humanoid robotics,'' in \emph{Adaptive motion of animals and machines}.\hskip 1em plus 0.5em minus 0.4em\relax Springer, 2006, pp. 261--280.

\bibitem{ijspeert2013dynamical}
A.~J. Ijspeert, J.~Nakanishi, H.~Hoffmann, P.~Pastor, and S.~Schaal, ``Dynamical movement primitives: learning attractor models for motor behaviors,'' \emph{Neural computation}, vol.~25, no.~2, pp. 328--373, 2013.

\bibitem{coumans2021}
E.~Coumans and Y.~Bai, ``Pybullet, a python module for physics simulation for games, robotics and machine learning,'' \url{http://pybullet.org}, 2016--2021.

\bibitem{movo}
\BIBentryALTinterwordspacing
Kinova, ``Kinova-movo.'' [Online]. Available: \url{https://docs.kinovarobotics.com/kinova-movo/Concepts/c_movo_hardware_overview.html}
\BIBentrySTDinterwordspacing

\bibitem{kirillov2023segment}
A.~Kirillov, E.~Mintun, N.~Ravi, H.~Mao, C.~Rolland, L.~Gustafson, T.~Xiao, S.~Whitehead, A.~C. Berg, W.-Y. Lo, \emph{et~al.}, ``Segment anything,'' \emph{arXiv preprint arXiv:2304.02643}, 2023.

\bibitem{wang2023rangedik}
Y.~Wang, P.~Praveena, D.~Rakita, and M.~Gleicher, ``Rangedik: An optimization-based robot motion generation method for ranged-goal tasks,'' \emph{arXiv preprint arXiv:2302.13935}, 2023.

\end{thebibliography}
}

\end{document}